\documentclass{article}

\usepackage{amsmath,amsfonts,amssymb,times,graphicx,natbib,algorithm,algorithmic,hyperref}

\usepackage[accepted]{whi2016}

\usepackage{mathtools}
\usepackage{eqnarray}
\usepackage{amsfonts,bbm,bm,courier}
\usepackage{booktabs}
\usepackage{textcomp}
\usepackage{gensymb}

\icmltitlerunning{Interpretable Machine Learning Models for the Digital Clock Drawing Test}

\begin{document} 

\twocolumn[
\icmltitle{Interpretable Machine Learning Models \\ 
           for the Digital Clock Drawing Test}

\icmlauthor{William Souillard-Mandar}{souillardmandar@csail.mit.edu}
\icmladdress{MIT Computer Science And Artificial Intelligence Laboratory ,
            32 Vassar Street, Cambridge, MA 02139 USA}
\icmlauthor{Randall Davis}{davis@csail.mit.edu}
\icmladdress{MIT Computer Science And Artificial Intelligence Laboratory ,
            32 Vassar Street, Cambridge, MA 02139 US}
\icmlauthor{Cynthia Rudin}{rudin@mit.edu}
\icmladdress{MIT Computer Science And Artificial Intelligence Laboratory ,
            32 Vassar Street, Cambridge, MA 02139 US}
\icmlauthor{Rhoda Au}{rhodaau@bu.edu}
\icmladdress{Boston University School of Medicine,
            72 E Concord St, Boston, MA 02118 US}
\icmlauthor{Dana L. Penney}{dana.l.penney@lahey.org}
\icmladdress{Lahey Hospital and Medical Center,
            41 Burlington Mall Road, Burlington, MA 01805 US}

\icmlkeywords{Machine Learning, ICML, WHI}

\vskip 0.3in
]

\begin{abstract} 
The Clock Drawing Test (CDT)  is a rapid, inexpensive, and popular neuropsychological screening tool for cognitive conditions.
The Digital Clock Drawing Test (dCDT) uses novel software to analyze data from a digitizing ballpoint pen that reports its position with considerable spatial and temporal precision, making possible the analysis of both the drawing process and final product. 
We developed methodology to analyze pen stroke data from these drawings, and computed a large collection of features which were then analyzed with a variety of machine learning techniques. The resulting scoring systems were designed to be more accurate than the systems currently used by clinicians, but just as interpretable and easy to use. The systems also allow us to quantify the tradeoff between accuracy and interpretability. We created automated versions of the CDT scoring systems currently used by clinicians, allowing us to benchmark our models, which indicated that our machine learning models substantially outperformed the existing scoring systems.
 
\end{abstract} 

\section{Background}
\label{background}
The Clock Drawing Test (CDT) - a simple pencil and paper test - has been used as a screening tool to differentiate normal individuals from those with cognitive impairment. The test takes less than two minutes,  is easily administered and inexpensive, and is deceptively simple: it asks subjects first to draw an analog clock-face showing 10 minutes after 11 (the {\em command clock}), then to copy a pre-drawn clock showing the same time (the {\em copy clock}). It has proven useful in helping to diagnose cognitive dysfunction associated with neurological disorders such as Alzheimer's disease, Parkinson's disease, and other dementias and conditions. \citep{freedman1994clock,ashendorf2013bostonwithrandy}. 
The CDT is often used by neuropsychologists, neurologists and primary care physicians as part of a general screening for cognitive change \citep{strub1985mental}.    

\begin{figure}[ht]
\begin{center}
\centerline{\includegraphics[width=\columnwidth]{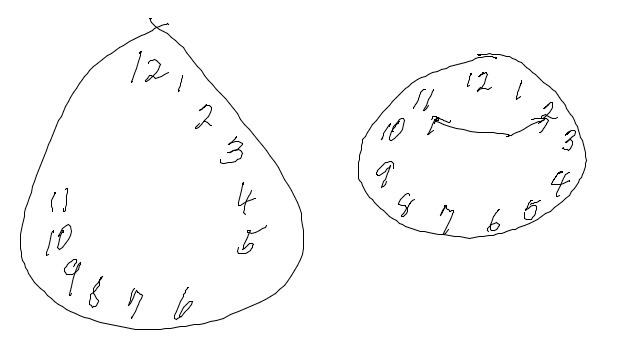}}
\caption{Example Alzheimer's Disease clock from our dataset.}
\label{fig:ExampleClockA}
\end{center}
\end{figure}

\begin{figure}[ht]
\begin{center}
\centerline{\includegraphics[width=\columnwidth]{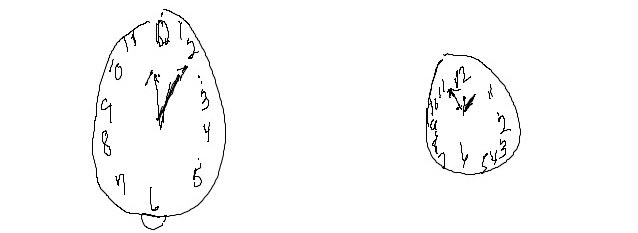}}
\caption{Example Parkinson's Disease clock from our dataset.}
\label{fig:ExampleClockP}
\end{center}
\end{figure}

For the past decade, neuropsychologists in our group have been administering the CDT using a commercially available digitizing ballpoint pen (the DP-201
from Anoto, Inc.) that records its
position on the page with considerable spatial ($\pm 0.005$ cm) and temporal (13ms)
accuracy, enabling the analysis of not only the end product -- the drawing -- but also the process that produced it, including all of the subject's movements and hesitations. 
The resulting test is called the digital Clock Drawing Test (dCDT). Figure \ref{fig:ExampleClockA} and Figure  \ref{fig:ExampleClockP} illustrate  clock drawings from a subject in the memory impairment group, and a subject diagnosed with Parkinson's disease, respectively.

\section{Existing Scoring Systems}

There are a variety of methods for scoring the CDT, varying in complexity and the types
of features they use. They often take the form of systems that add or subtract points based on features
of the clock, and often have the additional constraint that the $(n+1)^{th}$ feature matters only if the previous $n$ features have been satisfied, adding a higher level of complexity in understanding the resulting score. A threshold is then used to decide whether the test gives evidence of impairment. 

While the scoring system are typically short and understandable by a human, the features they attend to are often expressed in relatively vague terms, leading to potentially lower inter-rater reliability. For example, the Rouleau \citep{rouleau1992quantitative} scoring system, shown in Table \ref{table:Rouleau}, asks whether there are ``slight errors in the placement of the hands" and whether ``the clockface is present without gross distortion". 

\begin{table}[h]
\centering
\footnotesize
\begin{tabular}{ p{7cm} p{1cm} }
\toprule
maximum: 10 points \\
\toprule
1.	Integrity of the clockface (maximum: 2 points) \\
\midrule
\hspace*{1ex} 2: Present without gross distortion \\
\hspace*{1ex} 1: Incomplete or some distortion \\
\hspace*{1ex} 0: Absent or totally inappropriate \\
\midrule
2.	Presence and sequencing of the numbers \\ (maximum: 4 points) \\
\midrule
\hspace*{1ex} 4: All present in the right order and at most minimal \\
\hspace*{1ex}\phantom{4: } error in the spatial arrangement \\
\hspace*{1ex} 3: All present but errors in spatial arrangement \\
\hspace*{1ex} 2: Numbers missing or added but no gross distortions \\
\hspace*{1ex}\phantom{4: } of the remaining numbers \\
\hspace*{1ex}\phantom{2: } Numbers placed in counterclockwise direction \\
\hspace*{1ex}\phantom{2: } Numbers all present but gross distortion in spatial \\
\hspace*{1ex}\phantom{4: }  layout \\
\hspace*{1ex} 1: Missing or added numbers and gross spatial \\
\hspace*{1ex}\phantom{4: } distortions \\
\hspace*{1ex} 0: Absence or poor representation of numbers \\
\midrule
3.	Presence and placement of the hands \\ (maximum: 4 points) \\
\midrule
\hspace*{1ex} 4: Hands are in correct position and the size difference \\
\hspace*{1ex}\phantom{4: }  is respected \\
\hspace*{1ex} 3: Sight errors in the placement of the hands or no \\
\hspace*{1ex}\phantom{4: } representation of size difference between the hands \\
\hspace*{1ex} 2: Major errors in the placement of the hands (signif-
\hspace*{1ex}\phantom{4: } icantly out of course including 10 to 11) \\
\hspace*{1ex} 1: Only one hand or poor representation of two hands \\
\hspace*{1ex} 0: No hands or perseveration on hands \\
\bottomrule
\end{tabular}
\caption{\label{table:Rouleau} Original Rouleau scoring system \citep{rouleau1992quantitative}}
\end{table}
  
In order to benchmark our models for the dCDT against existing scoring systems, we needed to create automated versions of them so that we could apply them to our set of clocks. We did this for  seven of the most widely used existing scoring systems \citep{souillard2015learning} by specifying the computations to be done in enough detail that they could be expressed unambiguously in code. As one example, we translated ``slight errors in
the placement of the hands" to ``exactly two hands present AND at most one hand with a
pointing error of between $\epsilon_1$ and $\epsilon_2$ degrees", where the $\epsilon_i$
are thresholds to be optimized. We refer to these new models as {\em operationalized} scoring systems.

\section{An Interpretable Machine Learning Approach}

\subsection{Stroke-Classification and Feature Computation}

\begin{figure}[ht]
\begin{center}
\centerline{\includegraphics[width=\columnwidth/3]{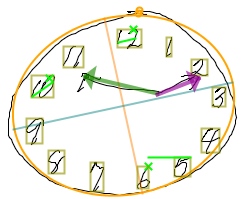}}
\caption{Classified command clock from Figure \ref{fig:ExampleClockA}}
\label{fig:ExampleClockClassified}
\end{center}
\end{figure}

The raw data from the pen is analyzed using novel software developed for this task
\citep{davis2014think,davis2014method,cohen2014digital}. An algorithm classifies the pen strokes as one or another of the clock drawing symbols (i.e. clockface, hands, digits, noise); stroke classification errors are easily corrected by human scorer using a simple drag-and-drop interface.
Figure \ref{fig:ExampleClockClassified} shows a screenshot of the system after the strokes in the command clock from Figure \ref{fig:ExampleClockA} have been classified.

Using these symbol-classified strokes, we compute a large collection of features from the test, measuring geometric and temporal properties in a single clock, both clocks, and differences between them. Example features include:
\begin{itemize}
\item The number of strokes; the total ink length; the time it took to draw; and the pen speed for various clock components; timing information is used to measure how quickly different parts of the clock were
drawn; latencies between components.
\item The length of the major and minor axis and eccentricity of the fitted ellipse; largest angular gaps in the clockface; distance and angular difference between starting and ending points of the clock face.
\item Digits that are missing or repeated; the height and width of digit bounding boxes.
\item Omissions or repetitions of hands; angular error from their correct angle; the hour hand to minute hand size ratio; the presence and direction of arrowheads.
\end{itemize}

We also selected a subset of our features  that we believe are both particularly understandable and that have values easily verifiable by clinicians. We expect, for example, that there would be wide agreement on whether a number is present, whether hands have arrowheads on them, whether there are easily noticeable noise
strokes, or if the total drawing time particularly high or low. We call this subset the \emph{Simplest Features}.

\subsection{Traditional Machine Learning}

We focused on three categories of cognitive impairment, for which we had a total of 453 tests:
memory impairment disorders (MID) consisting of Alzheimer's disease and amnestic mild cognitive
impairment (aMCI); vascular cognitive disorders (VCD) consisting of vascular dementia, mixed MCI
and vascular cognitive impairment; and Parkinson's disease (PD).
Our set of 406 healthy controls (HC) comes from people who have been longitudinally studied as participants in the Framingham Heart Study.

Our task is screening: we want to distinguish between
healthy and one of the three categories of cognitive impairment, as well as a group screening, distinguish between healthy and  all three conditions together.

We started our machine learning work by applying state-of-the-art machine learning methods to the set of all features. We generated classifiers using multiple machine learning methods, including CART
\citep{Breiman84}, C4.5 \citep{Quinlan93}, SVM with gaussian kernels \citep{Joachims/98c}, random
forests \citep{breiman2001random}, boosted decision trees \citep{friedman2001greedy}, and
regularized logistic regression \citep{fan2008liblinear}. We used stratified cross-validation to
divide the data into 5 folds to obtain training and testing sets. We further cross-validated
each training set into 5 folds to optimize the parameters of the algorithm using grid search
over a set of ranges. We chose to measure quality using area under the receiver operator characteristic curve (AUC) as a
single, concise statistic. 

We found that the AUC for best classifiers ranged from 0.88 to 0.93. We also ran our experiment on the subset of Simplest Features, and found that the AUC ranged from 0.82 to 0.83. Finally, we measured the performance of the operationalized scoring systems; the best ones ranged from 0.70 to 0.73. Complete results can be found in Table \ref{table:auc}.

\subsection{Human Interpretable Machine Learning}

\subsubsection{Definition of Interpretability}

To ensure that we produced models that can be
used and accepted in a clinical context, we obtained guidelines from clinicians. This led us to
focus on three components in defining complexity:
 
{\bf Computational complexity:} the models should be relatively easy to compute, requiring a small
  number of simple operations, similar to the existing manual scoring systems.  Those 
 systems have on average 8 to 15 rules, with each rule containing on
  average one or two features. We thus focus on models that use fewer than 20 features, and have
  a simple form.

{\bf Understandability:} the rationale for a decision made by the model should be
  easily understandable, so that the user can understand why the prediction was made and
  can easily explain it. Thus if several features are roughly equally useful in the model,
  the most understandable one should be used. 

{\bf Ease of feature measurement:} 
Features that can be easily understood and verified by eye should be prioritized; 
this lead to the creation of the Simplest Features subset mentioned above.

\subsubsection{Supersparse Linear Interpretable Models}

\begin{table*}[t]
\caption{Classification results for the screening task: 
distinguishing clinical group from Healthy Control. Each entry in the table shows the mean and standard deviation of the AUC of a machine learning algorithm across 5 folds}
\label{table:auc}
\centering
\footnotesize
\begin{tabular}{lcccccccc}
\toprule
Algorithm &Memory impairment disorders  & Vascular cognitive disorders & PD & All three       \\
&  vs$.$ HC &  vs$.$ HC &  vs$.$ HC&  vs$.$ HC\\
\midrule
Best operationalized scoring system             & 0.73 (0.08)     & 0.72 (0.09)       & 0.73 (0.09)      &  0.70 (0.06) \\
Best ML with simplest features               & 0.83 (0.06)    & 0.82 (0.07)       & 0.83 (0.08)      & 0.83 (0.07) \\
Best ML with all features           & 0.93 (0.09)     & 0.88 (0.11)       & 0.91 (0.11)      & 0.91 (0.09) \\
SLIM with simplest features                            		&  0.78 (0.08)     &  0.75 (0.05)       &  0.78 (0.07)      &  0.74 (0.05) \\
SLIM with all features                           		&  0.83 (0.09)     & 0.81 (0.13)       &  0.81 (0.10)      & 0.83 (0.09) \\
\bottomrule
\end{tabular}
\end{table*}

We use a recently developed framework, Supersparse Linear Interpretable Models
(SLIM) \citep{ustun2015supersparse}, designed to create sparse
linear models that have integer coefficients. The framework produces models that meet many of our interpretability goals. Integer coefficients allow for models that are more easily computable, have greater expository power, and have the same form as the scoring systems already in use;  hard constraints on the coefficients allow us to set a hard limit on the number of variables used in the model, thus reducing computational complexity.

To improve model understandability, we added feature preferences by introducing an understandability penalty that indicates which features would be preferred over others when their performance is similar.

Given a dataset of $N$ examples $D_N = \{(\boldsymbol{x}_i,y_i) \}_{i=1}^N$, where observation $\boldsymbol{x}_i\in \mathrm{R}^J$ and label $y_i\in\{-1,1\}$, and an extra element with value 1 is included within each $\mathbf{x}_i$ vector to act as the intercept term, we want to build models of the form $ \hat{y} =
\text{sign}(\boldsymbol{\lambda}^T\boldsymbol{x})$, where $\boldsymbol{\lambda} \subseteq
\mathbb{Z}^{J+1}$ is a vector of integer coefficients. The framework determines the
coefficients of the models by solving an optimization problem of the form:
$$\underset{\boldsymbol{\lambda}}{\text{min}} \qquad \text{Loss}(\boldsymbol{\lambda};D_N) + \cdot \Phi(\boldsymbol{\lambda}) $$ 
$$  \text{s.t.} \quad \boldsymbol{\lambda} \in \mathcal{L}.$$
The Loss function is:
$$
\text{Loss}(\boldsymbol{\lambda};D_N) = C_+\frac{1}{N}\sum_{i:y_i=1} \psi_i + C_- \frac{1}{N}\sum_{i:y_i=-1} \psi_i,
$$
where $\psi_i$ is 1 if an incorrect prediction is made. 
It penalizes misclassifications and allows to set relative costs for accuracy on the positive examples and accuracy on the negative examples by setting $C_+$ and $C_-$.

The interpretability penalty function
$\Phi(\boldsymbol{\lambda}) \colon \mathbb{R}^{J+1}\rightarrow \mathbb{R}$ is defined as
\begin{equation} \label{eq1}
\begin{split}
\Phi(\boldsymbol{\lambda}) & = \text{sparsity penalty + understandability penalty} \\
 & = C_0 \sum_{j=1}^{J}{\mathbf{1}[\boldsymbol{\lambda}_j \neq 0]} +  C_1 \sum_{j=1}^{J}{u_j \cdot \mathbf{1}[\boldsymbol{\lambda}_j \neq 0]}.
\end{split}
\end{equation}

The first term computes the  count of the number of nonzero features, encouraging the model to use fewer
features. The second term allows for the prioritization of certain features, helping to ensure that the most understandable features appear in the
model. In particular, we defined an understandability penalty $u_j$ for each feature $j$ by organizing our
features into trees such that the children of each feature are those it depends on.  For
instance ``total time to draw both clocks" has as children ``total time to draw command clock"
and ``total time to draw copy clock." The height of a given node is the number of nodes traversed from the top of the tree to the given node.
We define
$$ u_j = \text{height($j$)} \quad \forall j $$ 
which produces a bias toward simpler features, i.e., those lower in the tree.
The constants $C_0$ and $C_1$ trade off between sparsity and understandability.

Given the above formulation, we used stratified cross-validation to divide the data into 5 folds to obtain training and testing sets and further cross-validated each training set into 5 folds to optimize the parameters ($C_+$, $C_-$, $C_0$, $C_1$) using grid search. 
We ran our optimization problem
on the set of simplest features and all features, with a hard upper bound of 10 features, to keep the resulting models
interpretable. 

The resulting AUCs ranged from 0.74 to 0.78 and 0.81 to 0.83, respectively. While this is lower than the traditional machine learning methods, it still outperforms existing scoring systems and, yet remains equally interpretable. As one example, the SLIM model for Memory Impairment screening containing only 9 binary features, yet it achieves an AUC score of 0.78 (Table \ref{table:SLIMAH}).  
 
\begin{table*}[t]
\centering
\footnotesize
\begin{tabular}{ p{13cm} p{2cm} }
\toprule
PREDICT MEMORY IMPAIRMENT DISORDER IF SCORE $<$ 10 \\
\toprule
Command clock: \\
\midrule
1. All digits are present, not repeated, and in the correct angular order & +5 \\

2. Hour hand is present & +5  \\

3. All of the non-anchor digits are in the correct eighth & +1 \\

4. Crossed-out digits present & -3 \\

5. Two hands not present & -1 \\

6. More than 60 seconds to draw & -1 \\

7. Minute hand points to digit 10 & -6\\
\midrule
Copy clock: \\
\midrule

8. All of the non-anchor digits are in the correct eighth & +4 \\

9. Numbers are repeated & -3  \\
\bottomrule
\end{tabular}
\caption{\label{table:SLIMAH} Supersparse Linear Integer Model for screening of memory impairment disorders}
\end{table*}

\section{Conclusion}
The dCDT combined with machine learning techniques allows for a significantly better screening of cognitive conditions than the existing CDT scoring systems. Traditional machine learning methods have high accuracy, but by constraining our models with formats similar to existing scoring systems, we can still obtain a significant improvement in accuracy and remove any subjectivity, while maintaining the human interpretability of the models. 



\bibliography{Bib}
\bibliographystyle{icml2016}

\end{document}